# Procode: the Swiss Multilingual Solution for Automatic Coding and Recoding of Occupations and Economic Activities

Nenad Savić[1*], Nicolas Bovio[1], Fabian Gilbert[2] and Irina Guseva Canu[1]

*The corresponding author:

**Nenad Savic**

Centre for primary care and public health – Unisanté

Route de la Corniche 2

1066 Epalinges-Lausanne

nenad.savic@unisante.ch

[1] Centre for Primary Care and Public Health (Unisanté), University of Lausanne, Route de la Corniche 2, CH-1066 Epalinges-Lausanne, Switzerland

[2] Research Institute for Environmental and Occupational Health, 28 rue Roger Amsler, CS 74521, 49045 Angers, France

**Number of words**: 2980 (excluding abstract, acknowledgement, funding and references)

**ABSTRACT**

**Objective.** Epidemiological studies require data that are in alignment with the classifications established for occupations/economic activities. The classifications usually include hundreds of codes and titles. Manual coding of raw data may result in misclassification and be time consuming. The goal was to develop and test a web-tool, named Procode, for coding of free-texts against classifications and recoding between different classifications.

**Methods.** Three text classifiers, i.e. Complement Naïve Bayes (CNB), Support Vector Machine (SVM) and Random Forest Classifier (RFC), were investigated using a k-fold cross-validation. 30'000 free-texts with manually assigned classification codes of French classification of occupations (PCS) and French classification of activities (NAF) were available. For recoding, Procode integrated a workflow that converts codes of one classification to another according to existing crosswalks. Since this is a straightforward operation, only the recoding time was measured.

**Results.** Among the three investigated text classifiers, CNB resulted in the best performance, where the classifier predicted accurately 57-81% and 63-83% classification codes for PCS and NAF, respectively. SVM lead to somewhat lower results (by 1-2%), while RFC coded accurately up to 30% of the data. The coding operation required one minute per 10'000 records, while the recoding was faster, i.e. 5-10 seconds.

**Conclusion.** The algorithm integrated in Procode showed satisfactory performance, since the tool had to assign the right code by choosing between 500-700 different choices. Based on the results, the authors decided to implement CNB in Procode. In future, if another classifier shows a superior performance, an update will include the required modifications.

# INTRODUCTION

Occupation is an essential component of adult life and a major determinant of health and healthy ageing (GBD., 2018). Moreover, some occupational exposures also affect the offspring health (Adachi et al., 2019; Pape et al., 2020). The burden of occupation-related diseases in the global burden of diseases (GBD) is heavily underestimated, mainly due to a shortage of occupational exposure data (Loomis, 2020). Large occupational exposure databases and job exposure matrices (JEMs) exist and could be used to assign exposures to occupations in valuable occupational, industrial, and population cohorts worldwide (Fadel et al., 2020). However, the use of JEMs is limited because of the time-consuming and costly work necessary to link such exposure data to individual occupational histories of cohort participants. In fact, information of occupational histories is usually provided as free-text entry and cannot be easily translated into a usable format (Koeman et al., 2013; Peters, 2020). Therefore, standardized and validated methods to translate such free text information into usable formats (i.e. standardized occupational coding systems) are needed to enable the usage of all available data for epidemiological and public health purposes. Computer-based tools are a cost-efficient alternative to manual coding or recoding. Several computer-based tools have been developed during the last decade (De Matteis et al., 2017; Patel et al., 2012; Remen et al., 2018; Russ et al., 2016). Some interpret job descriptions automatically, while some provide assistance either to professional coders or directly to subjects from a study. However, with exception of CAPS-Canada, which is bilingual (English and French) and allows recording occupation and economic activity from one of seven International, Canadian and US classifications to another (Remen et al., 2018), all the other tools only cover national US (Patel et al., 2012; Russ et al., 2016) or UK (De Matteis et al., 2017) classifications, in English. This limits their use in other countries, in particular in Switzerland where studies and surveys are conducted in four official languages. Moreover, none of the existing tools can proceed with automatic coding and recording. We thus aimed at creating a tool, which like a Swiss knife would provide a universal solution to code free-text occupations and economic activities regardless the language. The tool is also conceptualized to support any classification.

**METHODOLOGY**

The software, named Procode, is a web-tool that supports both coding of free-text data and recoding between different classifications. For these two operations, the authors developed two different algorithms. The entire development and testing of the software was conducted in Python. The Django framework was used for the backend side, while its frontend was developed in React.JS (Foundation, 2020; Facebook, 2020). The tool allows for (re-)coding of either single free-text entries or files holding multiple records.

Table 1 lists the classifications and the data that was available to train the coding algorithm of Procode. This list is not definitive and the authors will consider other classifications when new data is available.

**Table 1.** Summary of classifications and available training datasets

| Type | Name | Description | Languages | Dataset size |
|---|---|---|---|---|
| Occupations (international) | ISCO 1988 | *International Standard Classification of Occupations* | English German | 2500[1] |
| | ISCO 2008 | | French Italian | [2] |
| Occupations (national) | NSP 1990 | *Swiss classification of occupations* | German French | 10'000[3] |
| | NSP 2000 | | Italian | [1] |
| | PCS 2010 | *French classification of occupations* | French | 30'000 |
| | NOC 2011 | *Canadian National Occupation Classification* | English French | [2] |
| | SOC 2010 | *Standard Occupational Classification (USA)* | English | [2] |
| Economic activities (national) | NAF 2008 | *French classification of activities* | French | 30'000 |

[1] Insufficient data size – coding performed through recoding
[2] No data – coding performed through recoding
[3] Free-texts available in English, French and Italian

As shown in Table 1, the majority of the free-text data was for PCS classification. Until the data size for the other classifications remains low, the coding for a given classification is performed through recoding. This means that, for example, when a user executes the coding of a free-text against ISCO 1988, Procode will first code it against PCS and then recode to ISCO 1988.

Table 2 lists the crosswalks that are available in Procode. Most of these crosswalks were already established elsewhere and available online. For NSP (Swiss classification of occupations), however, there were no crosswalks from or to other classifications. A database of a Swiss national cohort, including 2.5 million records, where each record was coded against both NSP (1990 and 2000) and ISCO 1988, was available (Bovio et al., 2020; Guseva Canu et al., 2019). The authors conducted a comparison to evaluate which NSP code figure most frequently for a given ISCO code and vice-versa. This evaluation resulted in crosswalks between NSP 1990, NSP 2000 and ISCO 1988.

**Table 2.** Summary of crosswalks in Procode

| From classification | To classification | Source |
|---|---|---|
| ISCO 1988 | ISCO 2008 | Internet |
|  | NSP 1990 | 2'500'000 records[1] |
|  | NSP 2000 | 2'500'000 records[1] |
| ISCO 2008 | ISCO 1988 | Internet |
|  | SOC 2010 | Internet |
| NSP 1990 | ISCO 1988 | 2'500'000 records[1] |
| NSP 2000 | ISCO 1988 | 2'500'000 records[1] |
| PCS 2003 | ISCO 1988 | Internet |
| NOC 2011 | ISCO 1988 | Internet |
| SOC 2010 | ISCO 2008 | Internet |

[1]Swiss national cohort with NSP and ISCO codes

**Coding**

Procode implemented a machine learning technique of a text classifier model. The study considered three different models, i.e. Complement Naïve Bayes (CNB), Support Vector Classifier (SVC) and Random Forest Classifier (RFC). Procode integrated the one with the best performance results; the testing step is explained later in the text. The text classifier model had to be trained using a large dataset with previously, manually, coded data. This means that the data had to include a variety of free-texts intended to designate occupations or economic activities with the corresponding classification codes assigned. Before the training step, the data was formatted, where the words with a low importance for predicting of the classification code were discarded. The method used for this was term-frequency–inverse document frequency (tf-idf). Moreover, a word may take a variety of forms (e.g., singular or plural). The goal was thus to reduce the number of these different forms. The words were stemmed to their root. For example, the words, such as "programmer", "programs", "programming", were converted to "program". Natural Language Toolkit (NLTK) of Python was used for this specific task (Bird et al., 2009).

In addition, Procode is designed to allow the end-users to code free-texts regardless if they are in English, German, French or Italian. The available training database (Table 1) was, however, only in French. Moreover, it is difficult to have data with free-texts in different languages. This is especially difficult for countries with one official language. The authors provided an alternative solution to mitigate this problem. When a user enters a free-text, Procode will translate it into the most frequent language of the training dataset. The text classifier of Procode will thus always train its algorithm using the most complete data from its database. Since the database is currently only in French, every free-text is translated in this language before being coded. This feature is enabled by using *translate* package of Python (Yin and Henter, 2020).

Workflow

Figure 1 illustrates an overall coding workflow of Procode. The end-user enters several input parameters required to code a free-text that designates an occupations or economic activity. In addition, the user must specify the classification for the coding operation, the desired precision of the classification code and the language of the entered free-text(s). Procode then filters the corresponding data needed to train the algorithm of the tool from its internal database. Next, the tool verifies if the training data exists in the language of the provided free-text. If no such data, Procode translates the free-texts to the language corresponding to the language of the training dataset. Only when the languages match, the tool executes the coding operation.

Following the coding task, Procode allows the users to contribute to improving the tool's performance by providing feedbacks regarding the prediction accuracy. The end-user may or may not agree with the obtained classification code. In case that the user disagrees with a given code, the tool asks the user to specify which of the codes established for a given classification matches best the entered free-text. Procode then merges the provided feedback information with the existing data. The users' feedbacks should improve the coding performance over time.

Performance test

Among the three considered text classifiers, CNB is designed for imbalanced datasets specifically. Imbalanced datasets are those where the number of records is distributed unequally over different classification codes. For example, International standardized classification of occupations, version 88 (ISCO 1988) contains almost 700 occupation codes and it would thus be difficult to obtain a dataset with even similar number of records for all the possibilities. The three classifiers are described in detail elsewhere and were imported from *sklearn* package of Python (Rennie et al., 2003; Pedregosa et al., 2011). The performance of the three classifiers was investigated and compared using a dataset of 30'000 individuals (Table 1) from Constances cohort (Zins et al., 2015), where occupations and economic activities were recorded and coded against French classification of occupations (PCS 2003; fr. *Professions et catégories socioprofessionnelles*) and French classification of activities (NAF 2008 ; fr. *Nomenclature d'activités française*), respectively (Goldberg et al., 2017). The data were used to train the considered classifiers to predict the classification codes of PCS, for occupations, and NAF, for economic activities. Procode's coding performance was tested using 5-k-fold cross-validation. As the dataset contained the records coded at different levels of precision, where, for example, code *10* is of level two, while code *111a* is of level four, for each level, the coding performance was tested separately.

**Recoding**

The recoding is a more straightforward operation. For a pair of classifications (e.g. ISCO 1988 and ISCO 2008), the corresponding crosswalk is defined by a list containing matching classification codes. Certain crosswalks are available online or upon request. The crosswalks that were implemented in Procode are given in Table 1.

Figure 2 shows the implemented recoding workflow. The end-user selects the starting classification and the one to which a recoding operation is conducted. Next, the user supplies a list of classification codes corresponding to the starting classification and chooses the language in which the resulting classification titles are displayed.

Recoding performance

Since inter-classification crosswalks determine the outputs of recoding, the authors only conducted a verification test. This was a simple comparison of recoding results in Procode with those in crosswalks. The goal was to ensure that the programming code implemented in the tool works as intended. In addition, the study evaluated the time needed for Procode to perform recoding of those 30'000 PCS codes obtained through the coding test. The codes were recoded to ISCO 1988, while counting the execution time.

## RESULTS

### Coding

The results in Tables 3-5 show the evaluated percentages of the accurately predicted classification codes using different text classifier models. As expected, the lower level of a given classification code (or precision), the better results were obtained. The study found that CNB resulted in the best accuracy (Table 3), where the model predicted accurate codes for 57-81% for PCS and 63-83% for NAF. Somewhat lower values (1-2%) were observed with SVC (Table 4). Finally, RFC (Table 5) showed much lower prediction performance, where, in the best case, it accurately predicted 31% classification codes.

**Table 3.** Results per different classification level for French classification of occupations (PCS) and French classification of activities (NAF) obtained using Complement Naïve Bayes (CNB).

| Level of precision[1] | PCS 2003 | | NAF 2008 | |
| :---: | :---: | :---: | :---: | :---: |
| | Number of different codes | Accuracy (%) | Number of different codes | Accuracy (%) |
| 1 | 8 | 81 | 21 | 83 |
| 2 | 24 | 73 | 88 | 79 |
| 3 | 42 | 70 | 272 | 68 |
| 4 | 497 | 57 | 615 | 66 |
| 5 | - | - | 732 | 63 |

[1]Usually corresponds to the number of digits in a classification code (e.g. ISCO 1988), but not always (e.g. PCS 2003)

**Table 4.** Results per different classification level for French classification of occupations (PCS) and French classification of activities (NAF) obtained using Support Vector Classifier (SVC).

| Level of precision | PCS 2003 | | NAF 2008 | |
| :---: | :---: | :---: | :---: | :---: |
| | Number of different codes | Accuracy (%) | Number of different codes | Accuracy (%) |
| 1 | 8 | 79 | 21 | 83 |
| 2 | 24 | 72 | 88 | 80 |
| 3 | 42 | 70 | 272 | 68 |
| 4 | 497 | 55 | 615 | 65 |
| 5 | - | - | 732 | 62 |

**Table 5.** Results per different classification level for French classification of occupations (PCS) and French classification of activities (NAF) obtained using Random Forest Classifier (RFC).

| Level of precision | PCS 2003 | | NAF 2008 | |
| --- | --- | --- | --- | --- |
| | Number of different codes | Accuracy (%) | Number of different codes | Accuracy (%) |
| 1 | 8 | 26 | 21 | 31 |
| 2 | 24 | 17 | 88 | 21 |
| 3 | 42 | 14 | 272 | 16 |
| 4 | 497 | 14 | 615 | 15 |
| 5 | - | - | 732 | 9 |

**Recoding**

Regarding the recoding performance of Procode, only its duration was evaluated. The recoding performance was tested using the same dataset that was used for coding. The coding test resulted in a file holding free-texts and assigned PCS codes. The PCS codes were then recoded to ISCO 1988. For a dataset of 30'000 records, this task took less than ten seconds. Moreover, Procode could have been used to further recode the resulting codes of ISCO 1988 into another classification, such as, for example, ISCO 2008 or ISCO 2019. This is particularly interesting for a classification without an established crosswalk that enables a direct recoding to a desired classification. The recoding to a target classification could thus be achieved through several intermediate steps.

**End-user tests**

Following a beta release of Procode, two end-users was asked to test the both tool's functions: coding and recoding. The users (i.e. testers) evaluated the performance (e.g. coding speed on large datasets), user-friendliness and coding accuracy. Around 10'000 occupational and economic activity, the first tester coded against PCS and NAF, respectively. As the manually assigned codes ("gold" standard) were available, the user calculated the accuracy of the predicted codes. Regarding the code's precision, 60-83% of PCS and 71-83% of NAF codes were assigned accurately. The user reported that the coding operation took two minutes.

The second tester ran two coding iterations on ~ 1000 free-texts against ISCO 88. Following the first iteration, the tester checked the accuracy of the assigned codes and provided feedbacks, where considered necessary. The tester reported improved accuracy for the second iteration, which took into account the provided feedback information. This tester also conducted a recoding between NSP 2000 and ISCO 88. Finally, both testers provided a positive feedback regarding the user-friendliness of Procode and a few suggestions on how it may be further improved.

**DISCUSSION**

Procode was primary developed to cope with issues related to (re-)coding of large datasets holding thousands of unstandardized data – titles of occupations/economic sectors that differ from those defined for a given classification. Over time, the tool is expected to become more precise; the more it is used, the more data are collected. This study found that CNB results in superior performance compared to SVC and RFC. Therefore, the authors implemented this text classifier in Procode. The use of CNB is, however, not definitive, since additional tests will be performed to compare its performance with other potential models (e.g. SVC or RFC) when new data becomes available.

Previously, other tools were developed to support (re-)coding of the occupational data, such as, for example, OSCAR or CAPS-Canada (De Matteis et al., 2017; Remen et al., 2018). These tools, however, require many interventions from the end-users. For example, following a free-text entry in OSCAR, the users are asked to decide between different outputs iteratively through a decision-tree model. Unlike OSCAR, CAPS-Canada outputs automatically the most probable outcomes for a searched entry. The outcomes are ordered according to assigned scores, which are calculated based on the presence of key words in different segments of a given classification. In other words, when a searched text does not contain any key word that appears in a classification, the tool would fail to provide any result. This means that most of these tools could be defined as simple search engines, as their algorithms do not allow for a dynamic grow of the data. Their prediction power thus remains unchanged over time. Procode, although currently based on a relatively small training set, implements a method that continuously improve its performance.

The authors assumed the manually assigned classification codes in Constances dataset to be correct and used them as the main resource to train the coding algorithm and evaluate its prediction performance (Ruiz et al., 2016). Regarding the top-level precision of the two classifications (i.e. 4 for PCS and 5 for NAF), correct codes were predicted for 57-66% of the tested data. The results were better for lower-level classification codes (i.e. those with less digits). For each data record, the coding algorithm had to choose between ~500 for PCS and ~700 for NAF different classification codes. This means that for the different classification codes, the training algorithm had, on average, less than 50 records (data size / N outcomes) per code to train its algorithm. Due to this relatively low number, the found results were considered satisfactory. Even if it assumed that 50 records were always different, this number could have not cover the whole "universe" of possible words used to designate a classification.

Procode, unlike the other tools, provides a multilingual environment and the free-text coding regardless the language of the training dataset. Although almost the entire training data is in French, the end-users are not prevented from coding the data in English, German and Italian. If the entered free-text does not match the language of the training data, it is translated using the Python's package mentioned in the methodology section. To have an insight in its performance, the authors created a list with a two hundred free-texts in English and French that designate different occupations. For both languages a separate coding was executed and the assigned classification codes (PCS and ISCO) matched in 95% of the cases. Of course, tests that are more comprehensive are required to prove this hypothesis. With the data increase – either by collecting more data from other sources or from the user feedbacks – the translation task is expected to become irrelevant, since the training dataset will be available in the language corresponding the free-texts.

Regarding recoding, the current version of Procode requires that the tool's administrator prepare the recoding pairs between two classifications. This is done either by obtaining a given crosswalk from an online source or by conducting a comparison analysis as the one done in this study. In future, the data increase might allow to defining the crosswalks between different classifications automatically. For a given classification, a well-trained algorithm could be used to predict classification codes that correspond to the titles of another classification. For example, the coding algorithm that is trained using a large dataset for classification A can be used to assign codes for the classification titles defined for classification B. This means that the titles of classification B are used as the free-text entries, for which codes of classification A are assigned. Since each title has a corresponding code, for each code in classification B, the most appropriate code in classification A would be assigned.

**Funding**

This work was supported jointly by the Swiss State Secretariat for Economic Affairs (SECO) and Federal office of Public Health (OFPH), grant N° 0947002262 "SACoProST".

**Acknowledgements**

The authors thank the Constances cohort consortium, particularly M. Zins and M. Goldberg from INSERM for providing data; C. Pilorget from Santé publique France and S.B. Petersen from the University of Copenhagen for providing crosswalks for some French and Danish classifications; and the Swiss federal statistic office for information regarding Swiss classifications. In addition, the authors would like to express their gratitude to Yara Shoman and Jose Paz from Unisanté for testing Procode software and providing useful inputs regarding the the manuscript.

**Competing Interests**

The authors declare no conflict of interest relating to the work presented in this article.

# Figures

**Figure 1.** Coding workflow

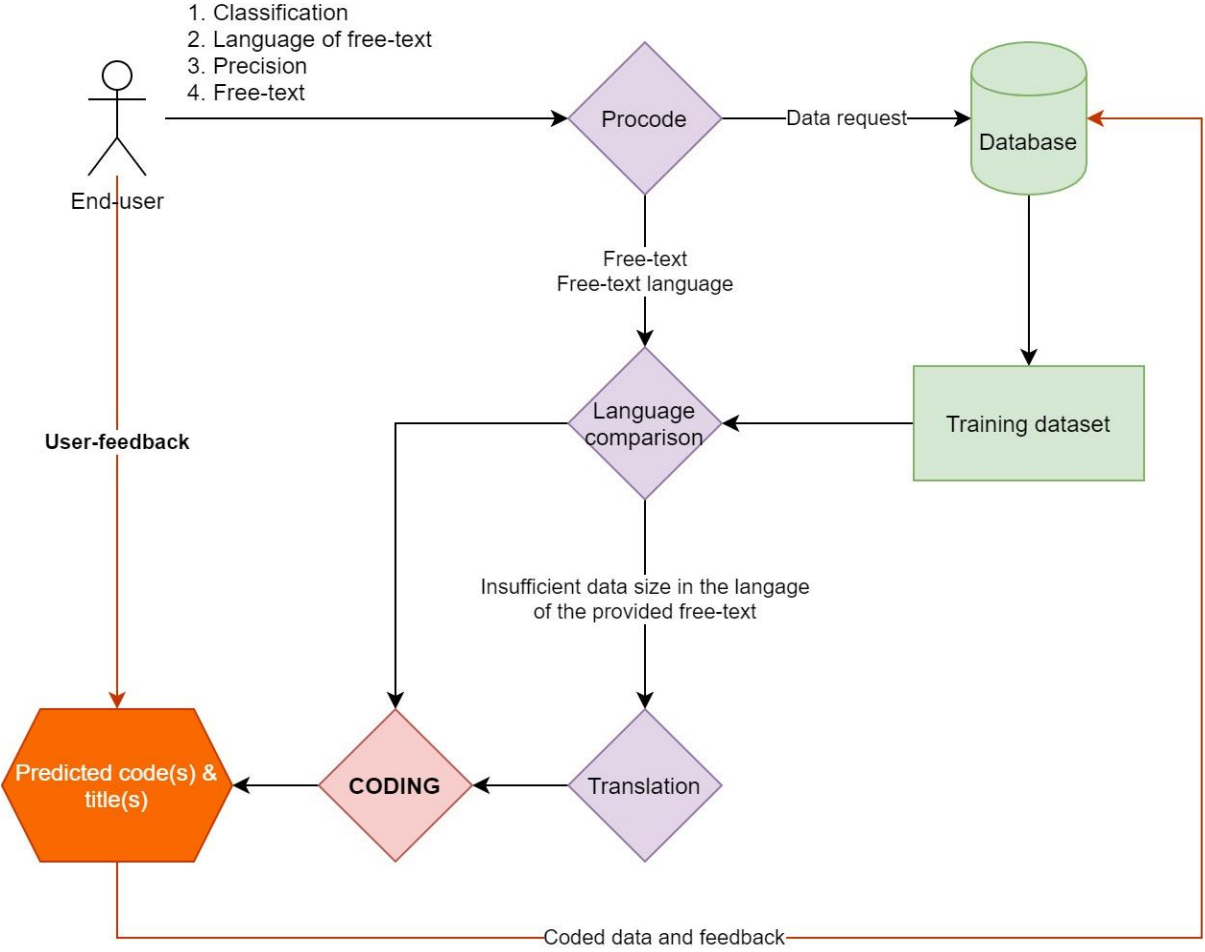

**Figure2.** Recoding workflow

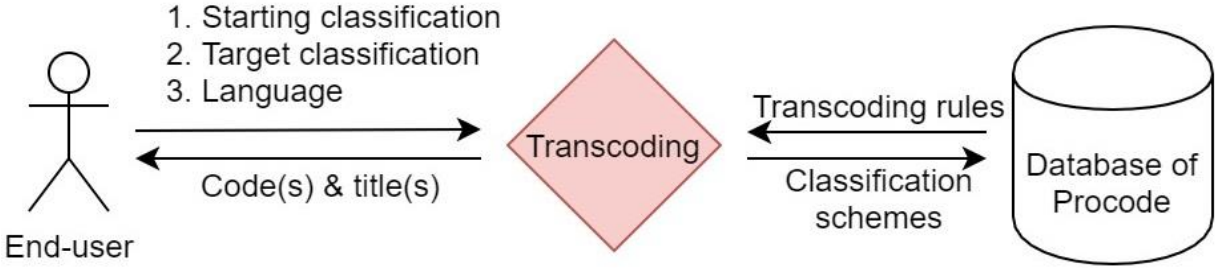